\documentclass[a4paper]{article}

\pdfoutput=1

    \usepackage[pdftex]{graphicx}

    \pdfcatalog {}

\usepackage{amsmath, amssymb, amsthm, mathrsfs, euscript}      
\usepackage{fancybox, fancyhdr}
\usepackage{ifthen}
\usepackage{version}

\makeatletter

\newcommand{\foot}[1]{%
  \edef\savecurrentlabel{\@currentlabel}%
                        \footnote{\hspace{1mm}%
                        \leftskip=-\parindent%
                        \ifthenelse{\value{footnote} < 10}%
			{\hangindent=6.7mm}{\hangindent=6.7mm}
                        #1}%
 \edef\@currentlabel{\savecurrentlabel}%
                       }
\newcommand{\pa}[1][]{\bigskip {\bf #1}}

\renewcommand{\paragraph}[1]{\pa\textbf{%
\mathversion{bold}#1\mathversion{normal}.}}

\newcommand{\expand}[2]{\begin{#1}#2\end{#1}}

\newcommand{\expandindex}[1]{\index{#1}}

\newcommand{\items}[1]{	      \expand{itemize}{\smallskip #1\smallskip}}
\newcommand{\enums}[1]{	      \expand{enumerate}{\smallskip #1\smallskip}}
\newcommand{\descriptions}[1]{{%
   \expand{description}{%
   \setlength{\leftskip}{7mm}%
   \smallskip #1\smallskip}%
}}

\newcommand{\dontsplit}[1] {\hbadness=10000\tolerance=10000%
                            \noindent\parbox[t]{\textwidth}{#1}
}

\newcommand{\form}[1] {\begin{equation}#1\end{equation}}
\newcommand{\formsplit}[1] {\form{\begin{split}#1\end{split}}}
\newcommand{\formskip}{\medskip}

\newcommand{\go}{\\[2.5mm]}
\newcommand{\eq}{\;=\;}
\newcommand{\fsequal}{&\eq}
\newcommand{\ap}{\;\approx\;}
\newcommand{\fsapprox}{&\ap}

\newcommand{\reference}[1]{\ref{#1} on page~\pageref{#1}}

\newcommand{\raggedparframe}[1]{

  \raggedbottom

  \pa

  \frame{\parbox{\textwidth}{#1}}

  \pa
  
  \flushbottom

}
\newcommand{\boldframe}[1]{{\raggedparframe{%
\hyphenpenalty10000%
\setlength{\leftskip}{7mm}%
\setlength{\rightskip}{7mm}%
\bf\pa #1\pa}}}

\newcommand{\margincite}[1]{\marginpar[\centering #1]{\centering #1}}

\renewcommand\section{\thispagestyle{plain}%
                      \@startsection {section}{1}{\z@}%
                                   {-3.5ex \@plus -1ex \@minus -.2ex}%
                                   {2.3ex \@plus.2ex}%
                                   {\normalfont\Large\bfseries%
				   }}

\makeatother

\usepackage{makeidx}

\newcommand{\setS}{\ensuremath{{\mathcal{S}}}}

\newcommand{\learn}{\ensuremath{{\EuScript L}}}

\newcommand{\kc}{Kolmogorov complexity}
\newcommand{\gaussian}{\ensuremath{{\mathcal{N}}}}
\newcommand{\thesisfigure}[1]{\begin{figure}[p]
    \sf\center \vspace{-.3cm}#1 \end{figure}
}

\newcounter{ExperimentCounter}

\thicklines

\fancypagestyle{plain}{\fancyhead{}}

\makeatletter
\renewcommand\subsection{\@startsection {section}{1}{\z@}%
                                   {-3.5ex \@plus -1ex \@minus -.2ex}%
                                   {2.3ex \@plus.2ex}%
                                   {\normalfont\Large\bfseries%
				   }}
\renewcommand{\l@section}{\l@subsection}
\makeatother

\begin{document}
\thispagestyle{empty}
\parindent=0cm

\title{A Short Introduction to Model Selection,\\Kolmogorov Complexity and\\ 
Minimum Description Length (MDL)\thanks{The Paradox of Overfitting \cite{Nannen:2003}, Chapter 1.}}
\author{Volker Nannen}
\renewcommand{\today}{}

\maketitle

\begin{abstract}

\noindent
The concept of overfitting in model selection is explained and demonstrated.
After providing some background information on information 
theory and Kolmogorov complexity, we provide a short explanation of 
Minimum Description Length and error minimization. We conclude with a 
discussion of the typical features of overfitting in model selection.
\end{abstract}

\thispagestyle{empty}
\thispagestyle{empty}

\subsection{The paradox of overfitting}

Machine learning is the branch of Artificial Intelligence that deals
with learning algorithms. Learning is a figurative description of what
in ordinary science is also known as \index{model!selection} model
selection and \index{generalization} generalization.  In computer
science a\index{model!definition} model is a set of
binary encoded values or strings, often the parameters of a function
or statistical distribution. Models that parameterize the same function
or distribution are called a family. Models of the same family are
usually indexed by the number of parameters involved. This number of
parameters is also called the degree or the dimension of the model.

\pa

To learn some real world phenomenon means to take some examples of the
phenomenon and to select a model that describes them well.  When such
a model can also be used to describe instances of the same phenomenon
that it was not trained on we say that it generalizes well or that it
has a small generalization error. The task of a learning algorithm is
to minimize this generalization error.

\pa

Classical learning algorithms did not allow for logical
dependencies~\cite{Minsky:1969} and were not very interesting to
Artificial Intelligence.  The advance of techniques like neural
networks with back-propagation in the 1980's and Bayesian networks in
the 1990's has changed this profoundly. With such techniques it is
possible to learn very complex relations. Learning algorithms are now
extensively used in applications like expert systems, computer vision
and language recognition. Machine learning has earned itself a central
position in Artificial Intelligence.

\pa

A serious problem of most of the common learning algorithms is
overfitting. \index{overfitting} Overfitting occurs when the models
describe the examples better and better but get worse and worse on
other instances of the same phenomenon. This can make the whole
learning process worthless. A good way to observe overfitting is to
split a number of examples in two, a training set, and a test set and
to train the models on the training set. Clearly, the higher the
degree of the model, the more information the model will contain about
the training set. But when we look at the generalization error of the
models on the test set, we will usually see that after an initial
phase of improvement the generalization error suddenly becomes
catastrophically bad.  To the uninitiated student this takes some
effort to accept since it apparently contradicts the basic empirical
truth that more information will not lead to worse predictions. We may
well call this {\em the paradox of overfitting\/}\index{overfitting!paradox of}.

\pa

It might seem at first that overfitting is a problem specific to
machine learning with its use of very complex models. And as some
model families suffer less from overfitting than others the ultimate
answer might be a model family that is entirely free from
overfitting. But overfitting is a very general problem that has been
known to statistics for a long time. And as overfitting is not the
only constraint on models it will not be solved by searching for model
families that are entirely free of it. Many families of models are
essential to their field because of speed, accuracy, easy to teach
mathematically, and other properties that are unlikely to be matched
by an equivalent family that is free from overfitting. As an example,
polynomials are used widely throughout all of science because of their
many algorithmic advantages. They suffer very badly from overfitting.
ARMA\index{model!ARMA} models are essential to signal processing and
are often used to model time series. They also suffer badly from
overfitting.  If we want to use the model with the best algorithmic
properties for our application we need a theory that can select the
best model from any arbitrary family.

\subsection{An example of overfitting}
\label{overfitting}
\index{overfitting!example}

Figure~\reference{overfitting-lorenz} gives a good example of
overfitting. The upper graph shows two curves in the two-dimensional
plane. One of the curves is a segment of the Lorenz
\index{Lorenz attractor} attractor, the other a 43-degree
\index{polynomials} polynomial. A Lorenz attractor is a complicated self
similar object. Here it is only
important because it is definitely not a polynomial and because its
curve is relatively smooth. Such a curve can be approximated well by a
polynomial.

\thesisfigure{

    \caption{An example of overfitting}
    \label{overfitting-lorenz}
    \center

    \includegraphics[height=.3\textheight,width=.9\textwidth]
 		    {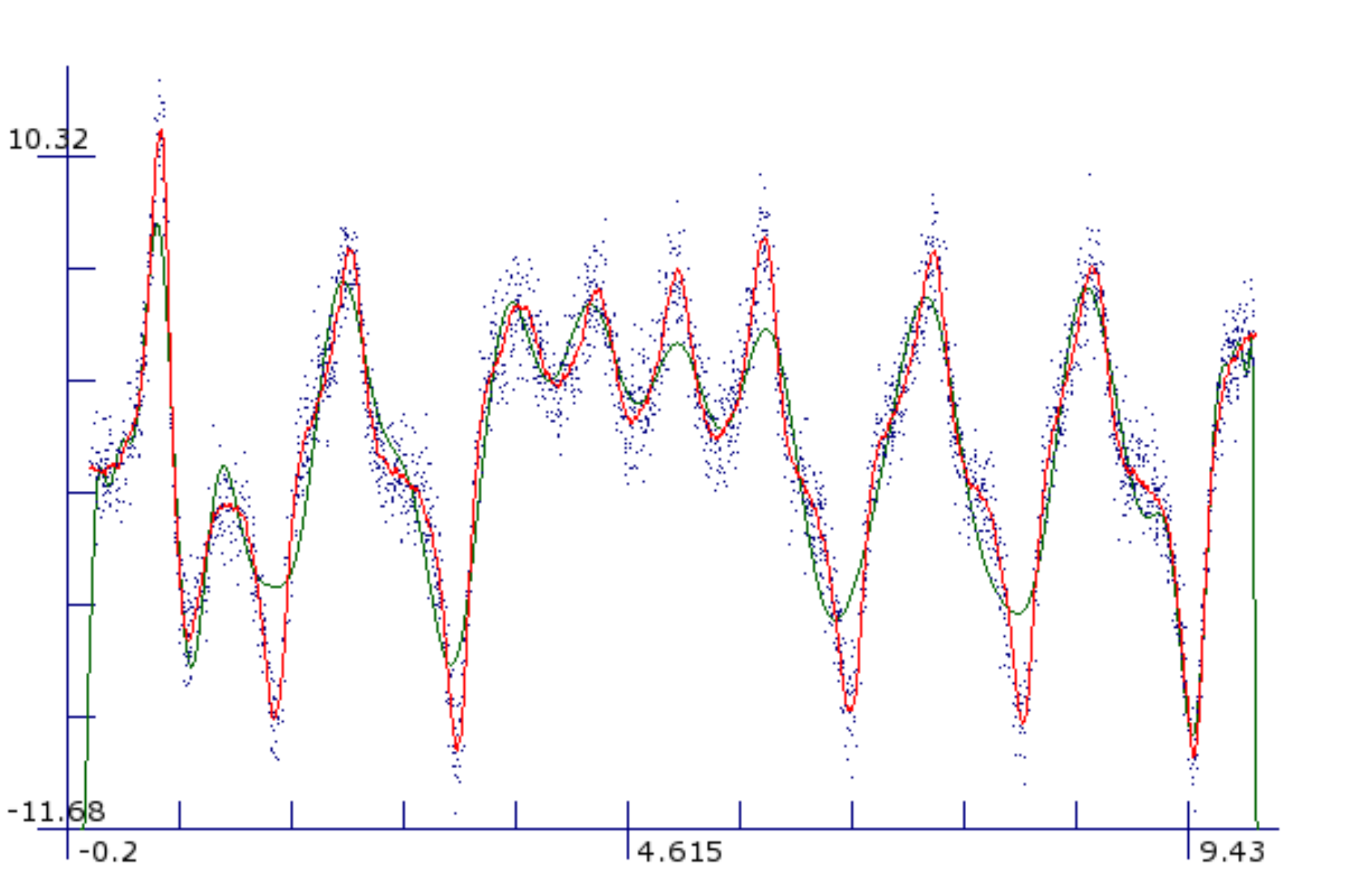}

    \medskip

    \parbox{\textwidth}{

  Lorenz attractor and the optimum 43-degree polynomial (the curve
  with smaller oscillations). The points are the 300 point training
  sample and the 3,000 point test sample.  Both samples are
  independently identically distributed.  The distribution over the
  $x$-axis is uniform over the support interval $[0,\,10]$. Along the
  $y$-axis, the deviation from the Lorenz attractor is Gaussian with
  variance $\sigma^2=1$.

    }

    \medskip

    \includegraphics[height=.3\textheight,width=.9\textwidth]
                    {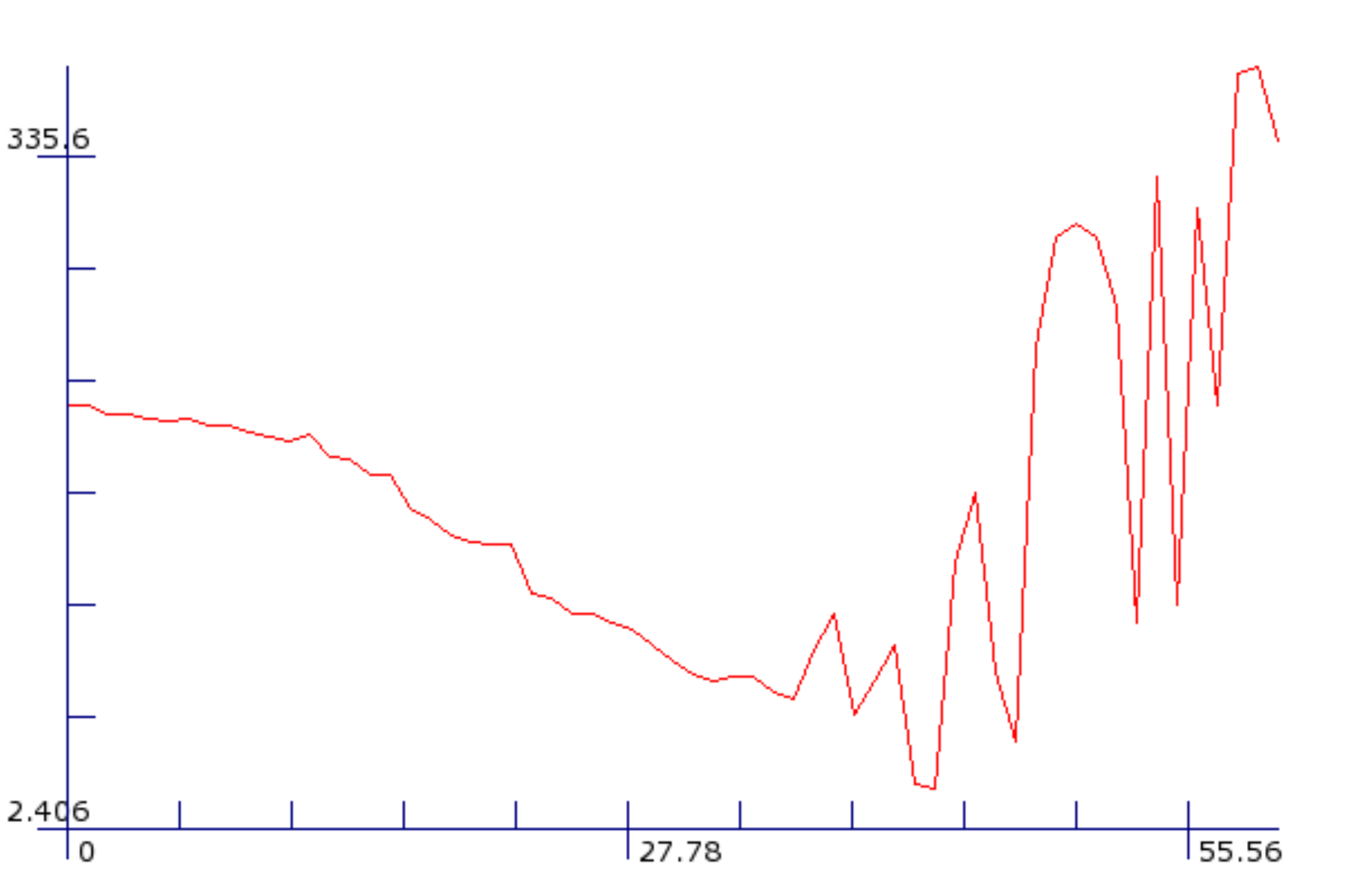}

    \medskip

    \parbox{\textwidth}{

  Generalization (mean squared error on the test set) analysis for
  polynomials of degree 0--60. The $x$-axis shows the degree of the
  polynomial. The $y$-axis shows the generalization error on the test
  sample. It has logarithmic scale.

  \medskip

  The first value on the left is the 0-degree polynomial.  It has a
  mean squared error of $\sigma^2=18$ on the test sample.  To the
  right of it the generalization error slowly decreases until it
  reaches a global minimum of $\sigma^2 =2.7$ at 43 degrees. After
  this the error shows a number of steep inclines and declines with
  local maxima that soon are much worse than the initial
  $\sigma^2=18$. 

    }

}

\pa

\dontsplit{

An $n$-degree polynomial is a function of the form
\\*
\form{
  f(x) \eq a_0 + a_1 x + a_2 x^2 + \dots + a_n x^n\;,\quad x\in\mathbb{R}
}

\formskip

with an $n+1$-dimensional parameter space $(a_0\dots
a_n)\in\mathbb{R}^{n+1}$. 

}

\pa

A polynomial is very easy to work with and
polynomials are used throughout science to model (or approximate)
other functions. If the other function has to be inferred from a
sample of points that witness that function, the problem is called a
regression \index{regression} problem.

\pa

Based on a small training sample that witnesses our Lorenz attractor
we search for a polynomial that optimally predicts future points that
follow the same distribution as the training sample---they witness the
same Lorenz attractor, the same noise along the $y$-axis and the same
distribution over the $x$-axis. Such a sample is called i.i.d.,
\index{i.i.d.} independently identically distributed. Here
the i.i.d.\@ assumption will be the only assumption about training
samples and samples that have to be predicted.

\pa

The Lorenz attractor in the graph is witnessed by 3,300 points. To
simulate the noise\index{noise} that is almost always polluting our
measurements, the points deviate from the curve of the attractor by a
small distance along the $y$-axis. They are uniformly distributed over
the interval $[0,\,10]$ of the $x$-axis and are randomly divided into
a 300 point training set and a 3,000 point test set. The interval
$[0,\,10]$ of the $x$-axis is called the \index{support} support.

\pa

The generalization analysis\index{generalization!analysis} in the
lower graph of Figure~\ref{overfitting-lorenz} shows what happens if
we approximate the 300 point training set by polynomials of rising
degree and measure the generalization error of these polynomials on
the 3,000 point test set. Of course, the more parameters we choose,
the better the polynomial will approximate the training set until
it eventually goes through every single point of the
training set. This is not shown in the graph. What is shown is the
generalization error on the 3,000 points of the i.i.d.
test set. The $x$-axis shows the degrees of
the polynomial and the $y$-axis shows the generalization
error. 

\pa

Starting on the left with a 0-degree polynomial (which is
nothing but the mean of the training set) we see that a polynomial
that approximates the training set well will also approximate the test
set. Slowly but surely, the more parameters the polynomial uses the
smaller the generalization error becomes. In the center of the graph,
at 43 degrees, the generalization error becomes almost zero.
But then something unexpected happens, at least in the eyes
of the uninitiated student. For polynomials of 44 degrees and higher
the error on the test set rises very fast and soon becomes much bigger
than the generalization error of even the 0-degree polynomial. Though
these high degree polynomials continue to improve on the training set,
they definitely do not approximate our Lorenz attractor any more. They
{\em overfit}. 

\subsection{The definition of a good model}
\label{good-model}
\index{model|(}

Before we can proceed with a more detailed analysis of model selection
we need to answer one important question: what exactly is a good
model.  And one popular belief which is persistent even among
professional statisticians has to be dismissed right from the
beginning: the model that will achieve the lowest generalization error
does {\em not\/} have to have the same degree or even be of the same
family as the model that originally produced the data.

\pa

To drive this idea home we use a simple 4-degree polynomial as a
source function. This polynomial is witnessed by a 100 point training
sample and a 3,000 point test sample. To simulate noise, the points
are polluted by a Gaussian distribution of variance $\sigma^2=1$ along
the $y$-axis. Along the $x$-axis they are uniformly distributed over
the support interval $[0,\,10]$. The graph of this example and the analysis of
the generalization error are shown in
Figure~\ref{overfitting-polynomial}. The generalization error
shows that a 4-degree polynomial has a comparatively high
generalization error. When trained on a sample of this size and noise
there is only a very low probability that a 4-degree polynomial will
ever show a satisfactory generalization error.  Depending on the
actual training sample the lowest generalization error is achieved for
polynomials from 6 to 8 degrees.

\thesisfigure{

    \caption{Defining a good model}
    \label{overfitting-polynomial}
    \center

    \medskip

    \includegraphics[height=.34\textheight,width=.9\textwidth]
        {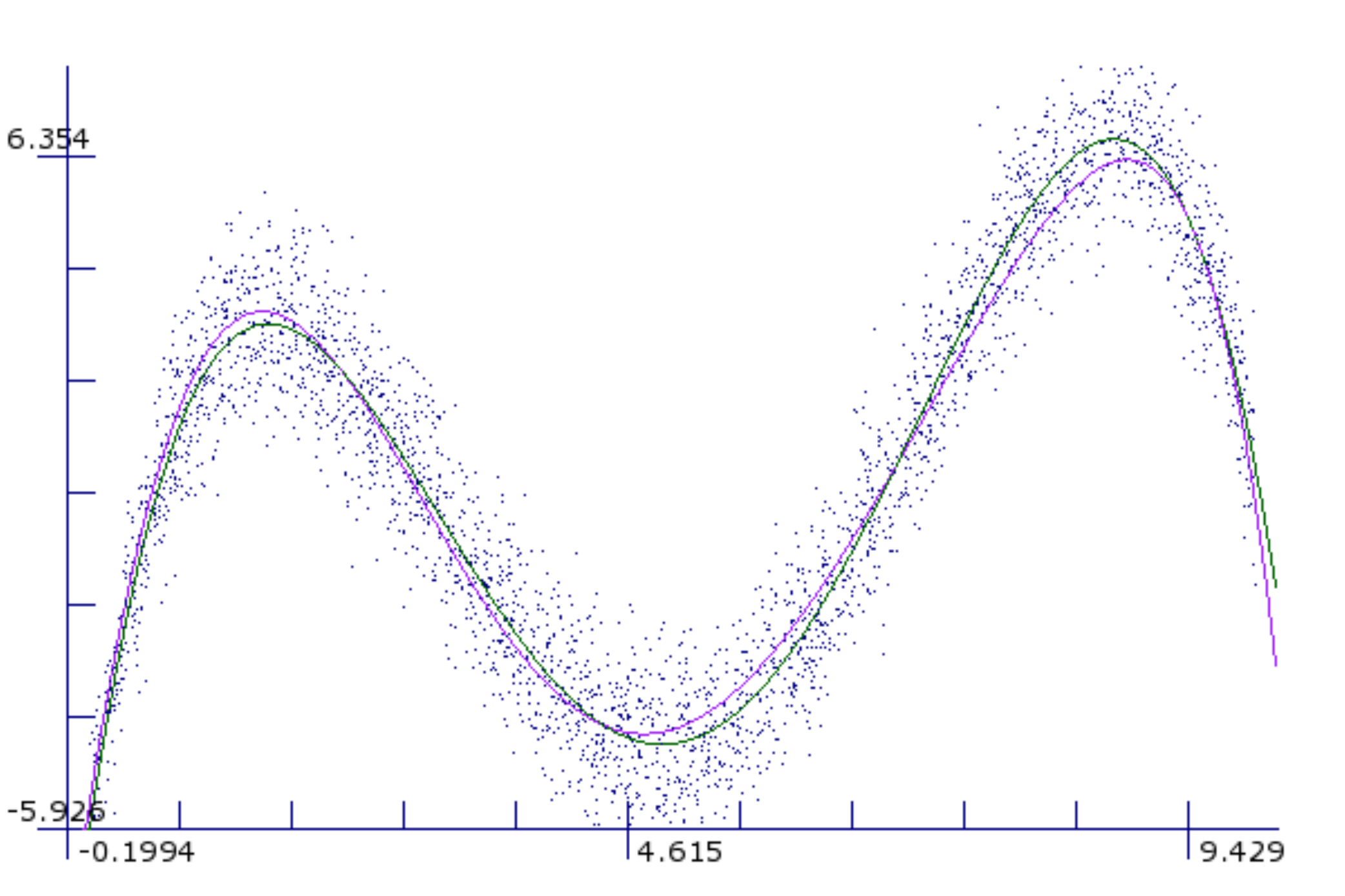}
    \parbox{\textwidth}{

    \medskip

  Original 4-degree polynomial (green), 100 point training sample,
  3,000 point test sample and 8-degree polynomial trained on the
  training sample (blue). In case you are reading a black and white
  print: the 8-degree polynomial lies above the 4-degree polynomial at
  the left peak and the middle valley and below the 4-degree
  polynomial at the right peak.

    }

    \medskip

    \includegraphics[height=.34\textheight,width=.9\textwidth]
        {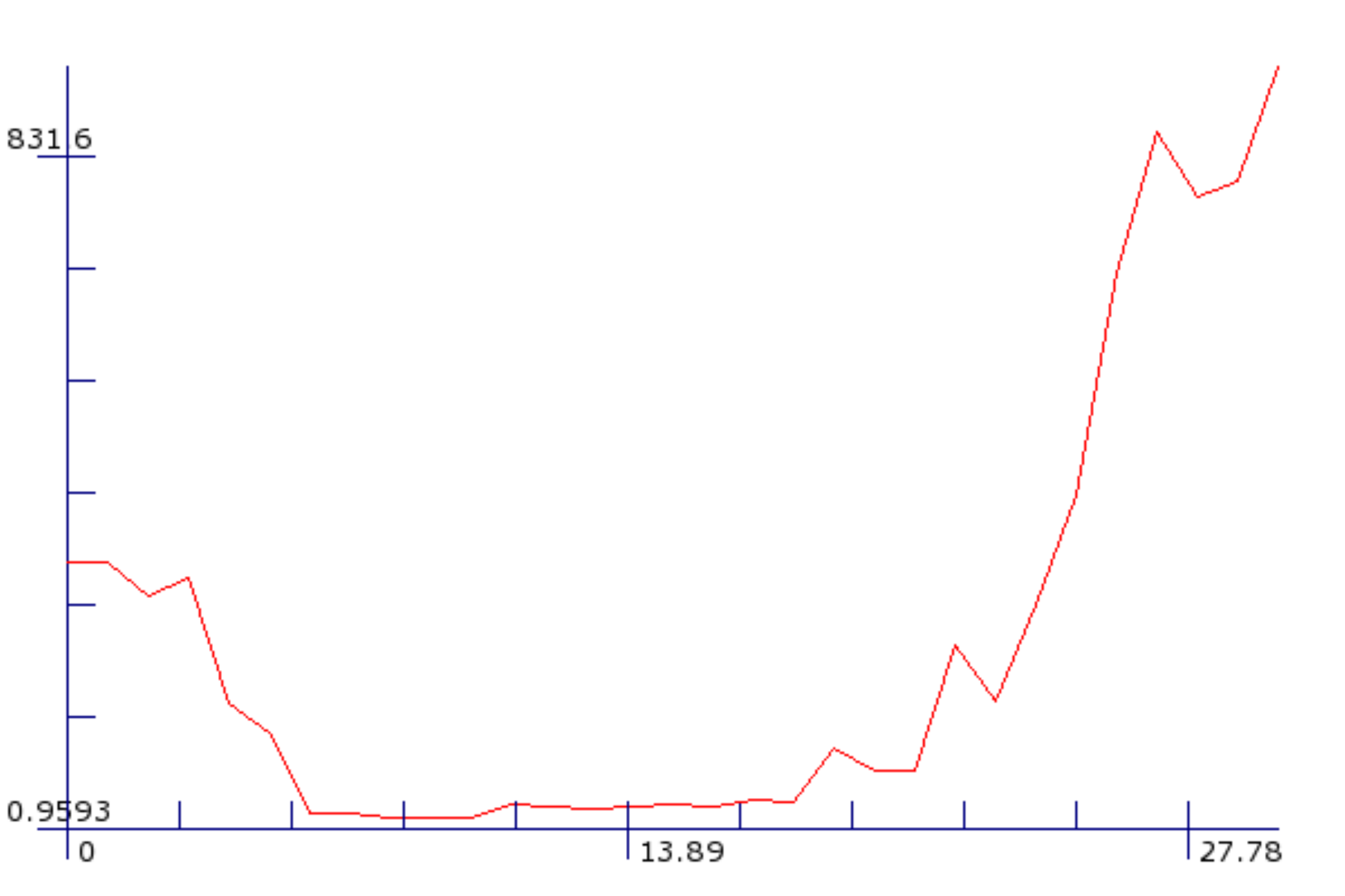}

    \medskip

    \parbox{\textwidth}{
 
  The analysis of the generalization error $\sigma^2$. A 0-degree
  polynomial achieves $\sigma^2=14$ and a 4-degree polynomial
  $\sigma^2=3.4$ on the test sample. All polynomials in the range
  6--18 degrees achieve $\sigma^2<1.3$ with a global minimum of
  $\sigma^2=1.04$ at 8 degrees. From 18 degrees onwards we witness
  overfitting. Different training samples of the same size might
  witness global minima for polynomials ranging from 6 to 8 degrees
  and overfitting may start from 10 degrees onwards. 4 degrees are
  always far worse than 6 degrees. The $y$-axis has logarithmic scale.

    }

}

\pa

This discrepancy is not biased by inaccurate algorithms. Neither can it
be dismissed as the result of an unfortunate selection of sample size,
noise and model family.  The same phenomenon can be witnessed for ARMA
models and many others under many different circumstances but
especially for small sample sizes.  In~\cite{Ruelle:1989} a number of
striking examples are given of rather innocent functions the output of
which cannot reasonably be approximated by any function of the same
family. Usually this happens when the output is very sensitive to
minimal changes in the parameters.  Still, the
attractor\index{attractor} of such a function can often be
parameterized surprisingly well by a 
very different family of functions\foot{
To add to the confusion, a function that accurately describes an
attractor is often advocated as the original function.  This can be
compared to confusing a fingerprint with a DNA string. Both are unique
identifiers of their bearer but only one contains his blueprint.}.


\pa

For most practical purposes a good model is a model that minimizes the
generalization error on future output of the process in question. But
in the absence of further output even this is a weak definition. We
might want to filter useful information from noise or to compress an
overly redundant file into a more convenient format, as is often the
case in video and audio applications. In this case we need to select a
model for which the data is most typical in the sense that the data is
a truly random member of the model and virtually indistinguishable
from all its other members, except for the noise. It implies that all
information that has been lost during filtering or lossy compression was
noise of a truly random nature. This definition of a good model is
entirely independent from a source and is known as {\em minimum
randomness deficiency}.\index{minimum randomness deficiency} It will
be discussed in more detail on page~\pageref{min-ran-def}.

\pa

\dontsplit{

We now have three definitions of a good model
\index{model!three definitions of good}:\label{three-definitions}

\enums{

  \setlength{\leftskip}{7mm}

  \item identifying family and degree of the original model for
        reconstruction purposes

  \item minimum generalization error for data prediction

  \item randomness deficiency for filters and lossy compression

}

}

\dontsplit{

We have already seen that a model of the same family and degree as the
original model does not necessarily minimize the generalization error.

\boldframe{

The important question is: can the randomness deficiency and the
generalization error be minimized by the same model selection
method?\index{model!selection}

}}

Such a general purpose method would simplify teaching and would enable
many more people to deal with the problems of model selection. A
general purpose method would also be very attractive to the embedded
systems industry. Embedded systems often hardwire algorithms and
cannot adapt them to specific needs. They have to be very economic
with time, space and energy consumption. An algorithm that can
effectively filter, compress and predict future data all at the same
time would indeed be very useful. But before this question can be
answered we have to introduce some mathematical theory.

\index{model|)}

\subsection{Information \& complexity theory}

This section provides a raw overview of the essential concepts. The 
interested reader is referred to the literature, especially the textbooks


\smallskip

\begin{tabbing}

blubla \= blubla \= \kill

\>  \bf\em Elements of Information Theory\\[1.2mm]
\>\>       by Thomas M. Cover and Joy A. Thomas,~\cite{Cover:1991}
        \index{entropy}\index{information theory}\\[5mm]

\> \bf\em Introduction to Kolmogorov Complexity and Its Applications\\[1.2mm]
\>\>        by Ming Li and Paul Vit\a'anyi,~\cite{Vitanyi:1997}
        \expandindex{\kc}

\end{tabbing}

\medskip

which cover the fields of information theory and \kc\ in depth and
with all the necessary rigor. They are well to read and require only a
minimum of prior knowledge.

\paragraph{\kc}%
\index{Kolmogorov, A.N.}%
\index{Chaitin, G.J.}%
\index{Solomonoff, R.}%
\expandindex{\kc}%
\expandindex{algorithmic complexity|see{\kc}}%
\expandindex{Turing complexity|see{\kc}}%
\index{universal Turing machine}%
\index{UTM|see{universal Turing machine}} 
The concept of \kc\ was developed independently and with different
motivation by Andrei N.~Kolmogorov \cite{Kolmogorov:1965}, Ray
Solomonoff \cite{Solomonoff:1964} and Gregory Chaitin
\cite{Chaitin:1966}, \cite{Chaitin:1969}.\foot{{\em \kc\/} is sometimes
also called {\em algorithmic complexity\/} and {\em Turing
complexity}. Though Kolmogorov was not the first one to formulate the
idea, he played the dominant role in the consolidation of the theory.}

\pa

The \kc\ $C(s)$\margincite{$C(\cdot)$} of any binary string
$s\in\{0,1\}^n$ is the length of the shortest computer program $s^*$
that can produce this string on the Universal Turing Machine
UTM\margincite{UTM} and then halt. In other words, on the UTM $C(s)$
bits of information are needed to encode $s$.  The UTM is not a real
computer but an imaginary reference machine. We don't need the
specific details of the UTM. As every Turing machine can be
implemented on every other one, the minimum length of a program on one
machine will only add a constant to the minimum length of the program
on every other machine. This constant is the length of the
implementation of the first machine on the other machine and is
independent of the string in question.  This was first observed in
1964 by Ray Solomonoff.

\pa

Experience has shown that every attempt to construct a theoretical
model of computation that is more powerful than the Turing machine has
come up with something that is at the most just as strong as the
Turing machine. This has been codified in 1936 by Alonzo Church as
\index{Church's Thesis} Church's Thesis: the class of algorithmically 
computable numerical functions coincides with the class of partial
recursive functions.  Everything we can compute we can compute by a
Turing machine and what we cannot compute by a Turing machine we
cannot compute at all. This said, we can use \kc\ as a universal
measure that will assign the same value to any sequence of bits
regardless of the model of computation, within the bounds of an
additive constant.

\paragraph{Incomputability of \kc}
\kc\ is not computable. It is nevertheless essential for proving
existence and bounds for weaker notions of complexity. The fact that
\kc\ cannot be computed stems from the fact that we cannot compute
the output of every program. More fundamentally, no algorithm is
possible that can predict of every program if it will ever halt, as
has been shown by Alan Turing in his famous work on the halting
problem~\cite{Turing:1936}. No computer program is possible that, when
given any other computer program as input, will always output
\verb_true_ if that program will eventually halt and \verb_false_ if
it will not. Even if we have a short program that outputs our string
and that seems to be a good candidate for being the shortest such
program, there is always a number of shorter programs of which we do
not know if they will ever halt and with what output.

\paragraph{Plain versus prefix complexity}
Turing's original model of computation included special delimiters that
marked the end of an input string. This has resulted in two brands of
\kc: 

\descriptions{

  \item[plain \kc:]\expandindex{\kc!plain}\index{C@$C(\cdot)$}
      the\margincite{$C(\cdot)$} length $C(s)$ of the shortest binary
      string that is delimited by special marks and that can compute
      $x$ on the UTM and then halt.

  \item[prefix \kc:]\expandindex{\kc!prefix}\index{K@$K(\cdot)$}
       the\margincite{$K(\cdot)$} length $K(s)$ of the shortest binary
       string that is {\em self-delimiting\/}~\cite{Vitanyi:1997} and
       that can compute $x$ on the UTM and then halt.

}

The difference between the two is logarithmic in $C(s)$: the number of
extra bits that are needed to delimit the input string. While plain
\kc\ integrates neatly with the Turing model of computation,
prefix \kc\ has a number of desirable mathematical
characteristics that make it a more coherent theory. The individual
advantages and disadvantages are described in~\cite{Vitanyi:1997}.
Which one is actually used is a matter of convenience. We will mostly
use the prefix complexity $K(s)$.

\paragraph{Individual randomness}%
\index{individual randomness}
A. N. Kolmogorov was interested in \kc\ to define the individual
randomness of an object.  When $s$ has no computable regularity it
cannot be encoded by a program shorter than $s$. Such a string is
truly random and its \kc\ is the length of the string itself plus the
commando {\sf print}\foot{Plus a logarithmic term if we use prefix
complexity}. And indeed, strings with a \kc\ close to their actual
length satisfy all known tests of randomness. A regular string, on the
other hand, can be computed by a program much shorter than the string
itself.  But the overwhelming majority of all strings of any length
are random and for a string picked at random chances are exponentially
small that its \kc\ will be significantly smaller than its actual
length.

\pa

This can easily be shown. For any given integer $n$ there are exactly
$2^n$ binary strings of that length and $2^n-1$ strings that are
shorter than $n$: one empty string, $2^1$ strings of length one, $2^2$
of length two and so forth. Even if all strings shorter than $n$ would
produce a string of length $n$ on the UTM we would still be one string
short of assigning a $C(s)<n$ to every single one of our $2^n$
strings. And if we want to assign a $C(s)<n-1$ we can maximally do so
for $2^{n-1}-1$ strings. And for $C(s)<n-10$ we can only do so for
$2^{n-10}-1$ strings which is less than $0.1\%$ of all our strings.  Even
under optimal circumstances we will never find a $C(s)<n-c$ for more
than $\frac{1}{2^c}$ of our strings.

\paragraph{Conditional \kc}%
\expandindex{\kc!conditional}%
\index{K@$K(\cdot"|\cdot)$}
The conditional \kc\ $K(s|a)$\margincite{$K(\cdot|\cdot)$} is defined
as the shortest program that can output $s$ on the UTM if the input
string $a$ is given on an auxiliary tape. $K(s)$ is the special case
$K(s|\epsilon)$ where the auxiliary tape is empty. 

%
%
%

\paragraph{The universal distribution}%
\index{universal distribution}%
\index{Bayes' rule}%
\index{Solomonoff, R.}
When Ray Solomonoff first developed \kc\ in 1964 he intended it to
define a universal distribution over all possible objects. His
original approach dealt with a specific problem of Bayes' rule, the
unknown prior distribution. \index{distribution!prior}
\index{prior|see{distribution}} Bayes' rule can be used to calculate
$P(m|s)$, the probability for a probabilistic model to have generated
the sample $s$, given $s$. It is very simple. $P(s|m)$, the
probability that the sample will occur given the model, is multiplied
by the unconditional probability that the model will apply at all,
$P(m)$. This is divided by the unconditional probability of the sample
$P(s)$. The unconditional probability of the model is called the prior
distribution and the probability that the model will have generated
the data is called the posterior distribution.
\index{posterior|see{distribution}} \index{distribution!posterior}

\form{
  P(m|s) \eq \frac{P(s|m)\;P(m)}{P(s)}
}

\pa

Bayes' rule can easily be derived from the definition of conditional
probability:

\form{
  P(m|s) \eq \frac{P(m,s)}{P(s)}
}
and
\form{
  P(s|m) \eq \frac{P(m,s)}{P(m)}
}

\pa

The big and obvious problem with Bayes' rule is that we usually have
no idea what the prior distribution $P(m)$ should be. Solomonoff suggested
that if the true prior distribution is unknown the best assumption
would be the universal distribution $2^{-K(m)}$ where $K(m)$ is the
\expandindex{\kc!prefix} prefix \kc\ of the
model\foot{\expandindex{\kc!plain}Originally Solomonoff used the plain
\kc\ $C(\cdot)$. This resulted in an improper distribution $2^{-C(m)}$
that tends to infinity. Only in 1974 L.A. Levin introduced prefix
complexity to solve this particular problem, and thereby many other
problems as well~\cite{Levin:1974}.}.  This is nothing but a modern
codification of the age old principle that is wildly known under the
name of \index{Occam's razor} Occam's razor: the simplest explanation
is the most likely one to be true.

\paragraph{Entropy}%
\index{Shannon, C.E.}%
\index{entropy}%
\index{code length}
Claude Shannon~\cite{Shannon:1948} developed information
theory\index{information theory} in the late 1940's. He was concerned
with the optimum code length that could be given to different binary
words $w$ of a source string $s$. Obviously, assigning a short code
length to low frequency words or a long code length to high frequency
words is a waste of resources. Suppose we draw a word $w$ from our
source string $s$ uniformly at random. Then the probability $p(w)$ is
equal to the frequency of $w$ in $s$.  Shannon found that the optimum
overall code length for $s$ was achieved when assigning to each word
$w$ a code of length $-\log p(w)$. Shannon attributed the original
idea to R.M. Fano and hence this code is called the
Shannon-Fano\index{Shannon-Fano code} code.  When using such an
optimal code, the average code length of the words of $s$ can be
reduced to

\form{
  H(s) \eq -\sum_{w\in s} p(w)\log p(w)
}

\formskip

where $H(s)$\margincite{$H(\cdot)$}\index{H@$H(\cdot)$} is called the
entropy of the set $s$. When $s$ is finite and we assign a code of
length $-\log p(w)$ to each of the $n$ words of $s$, the total code
length is

\form{
  -\sum_{w\in s}\log p(w) \eq n\, H(s)
}

\pa

Let $s$ be the outcome of some random process $W$ that produces the
words $w\in s$ sequentially and independently, each with some known
probability \mbox{$p(W=w)>0$}.  $K(s|W)$ is the \kc\ of $s$ given $W$.
Because the Shannon-Fano code is optimal, the probability that
$K(s|W)$ is significantly less than $n H(W)$ is exponentially
small. This makes the negative log likelihood of $s$ given $W$ a good
estimator of $K(s|W)$:

\formsplit{
  K(s|W) 
\fsapprox
  n\,H(W)
\go\fsapprox
  \sum_{w\in s}\log p(w|W)
\go\fsequal
  -\log p(s|W)
}

\paragraph{Relative entropy}%
\label{relative-entropy}%
\index{entropy!relative|see{Kullback Leibler distance}}%
\index{Kullback Leibler distance}%
\index{D@$D(\cdot"|"|\cdot)$}
The relative entropy $D(p||q)$\margincite{$D(\cdot||\cdot)$} tells us
what happens when we use the wrong probability to encode our source
string $s$.  If $p(w)$ is the true distribution over the words of $s$
but we use $q(w)$ to encode them, we end up with an average of
$H(p)+D(p||q)$ bits per word. $D(p||q)$ is also called the Kullback
Leibler distance between the two probability mass functions $p$ and
$q$. It is defined as

\form{
  D(p||q) \eq \sum_{w\in s}p(w)\log\frac{p(w)}{q(w)}
}

\paragraph{Fisher information}%
\label{fisher}%
\index{Fisher information}
Fisher information was introduced into statistics some 20 years before
C. Shannon introduced information theory~\cite{Fisher:1925}. But it
was not well understood without it. Fisher information is the variance
of the score $V$ of the continuous parameter space of our models
$m_k$. This needs some explanation. At the beginning of this  we
defined models as binary strings that discretize the parameter space
of some function or probability distribution. For the purpose of
Fisher information we have to temporarily treat a model $m_k$ as a
vector in $\mathbb{R}^k$. And we only consider models where for all
samples $s$ the mapping $f_s(m_k)$ defined by $f_s(m_k)=p(s|m_k)$ is
differentiable. Then the score $V$ can be defined as

\formsplit{
  V 
\fsequal
  \frac{\partial}{\partial\, m_k}\; \ln p(s|m_k)
\go\fsequal
  \frac{\frac{\partial}{\partial\, m_k}\;p(s|m_k)}{p(s|m_k)}
}

\pa

The score $V$ is the partial derivative of $\ln\,p(s|m_k)$, a term we
are already familiar with. The Fisher information
$J(m_k)$\index{J@$J(\cdot)$}\margincite{$J(\cdot)$} is

\form{ 
  J(m_k) \eq E_{m_k}\left[\frac{\partial}{\partial\,m_k}\;\ln
  p(s|m_k) \right]^2 
}

\pa

Intuitively, a high Fisher information means that slight changes to
the parameters will have a great effect on $p(s|m_k)$.  If $J(m_k)$ is
high we must calculate $p(s|m_k)$ to a high
precision\index{precision}\index{rounding}. Conversely, if $J(m_k)$ is
low, we may round $p(s|m_k)$ to a low precision.

\paragraph{\kc\ of sets}%
\expandindex{\kc!of sets}
The \kc\ of a set of strings \setS\ is the length of the shortest
program that can output the members of \setS\ on the UTM and then
halt.  If one is to approximate some string~$s$ with $\alpha < K(s)$
bits then the best one can do is to compute the smallest set $\setS$
with $K(\setS)\le\alpha$ that includes $s$. Once we have some
$\setS\ni s$ we need at most $\log|\setS|$ additional bits to
compute~$s$. This set \setS\ is defined by the Kolmogorov structure
function\index{Kolmogorov structure
function}\margincite{$h_s(\cdot)$}\index{h@$h_s(\cdot)$}

\form{
  h_s(\alpha) \eq \operatornamewithlimits{min}_\setS\big[\log
     |\setS| : \setS\ni s,\; K(\setS)\le\alpha\big]
}

\formskip

which has many interesting features. The function $h_s(\alpha)+\alpha$
is non increasing and never falls below the line $K(s)+O(1)$ but can
assume any form within these constraints. It should be evident that

\form{\label{ksf-set}
  h_s(\alpha) \;\ge\; K(s) - K(\setS)
}

\paragraph{\kc\ of distributions}%
\expandindex{\kc!of distributions}
The Kolmogorov structure function is not confined to finite sets.  If
we generalize $h_s(\alpha)$ to probabilistic models $m_p$ that define
distributions over $\mathbb R$ and if we let~$s$ describe a real
number, we obtain

\form{\label{ksf-distribution}
  h_s(\alpha) \eq \operatornamewithlimits{min}_{m_p}\big[-\log p(s|m_p)
     : p(s|m_p)>0,\; K(m_p)\le\alpha\big]
}

\formskip

where $-\log p(s|m_p)$ is the number of bits we need to encode~$s$
with a code that is optimal for the distribution defined
by~$m_p$. Henceforth we will write $m_p$
when the model\index{model} defines a probability distribution and
$m_k$
with $k\in\mathbb{N}$ when the model defines a probability
distribution that has $k$ parameters.  A set \setS\ can be viewed
as a special case of $m_p$, a uniform distribution with

\form{
 p(s|m_p) \eq \begin{cases}
   \; \frac{1}{|\setS|}   &   \text{if }\; s\in\setS 
\go
   \; 0              &   \text{if }\; s\not\in\setS
 \end{cases}
}

\paragraph{Minimum randomness deficiency}%
\label{min-ran-def}%
\index{minimum randomness deficiency}%
\index{randomness deficiency}
The randomness deficiency of a string~$s$ with regard to a model $m_p$
is defined as\margincite{$\delta(\cdot|m_p)$}\index{d@$\delta(\cdot"|m_p)$}

\form{
  \delta(s|m_p) \eq -\log p(s|m_p) - K(\,s|m_p,\, K(m_p)\,)
}

\formskip

for $p(s)>0$, and $\infty$ otherwise. This is a generalization of the
definition given in~\cite{Vitanyi:2002} where models are finite
sets. If $\delta(s|m_p)$ is small, then~$s$ may be considered a {\em
typical\/} or {\em low profile\/} instance of the distribution. $s$
satisfies {\em all\/} properties of low \kc\ that hold with high
probability for the support set of~$m_p$. This would not be the case
if $s$ would be exactly identical to the mean, first momentum or any
other special characteristic of $m_p$.

\pa

Randomness deficiency is a key concept to any application of \kc. As
we saw earlier, \kc\ and conditional \kc\ are not computable. We can
never claim that a particular string $s$ does have a conditional \kc\

\form{
  K(s|m_p) \ap -\log p(s|m_p)
}

\formskip

The technical term that defines all those strings that do satisfy this
approximation is {\em
typicality\index{typicality}\margincite{typicality}}, defined as a
small randomness deficiency $\delta(s|m_p)$.

\pa

Minimum randomness deficiency turns out to be important for lossy data
compression.  A compressed string of minimum randomness deficiency is
the most difficult one to distinguish from the original string. The
best lossy compression that uses a maximum of $\alpha$ bits is defined
by the minimum randomness deficiency
function

\margincite{$\beta_s(\cdot)$}\index{b@$\beta_s(\cdot)$}

\form{
  \beta_s(\alpha) \eq \operatornamewithlimits{min}_{m_p}\big[\delta(s|m_p):
  p(s|m_p)>0,\; K(m_p)\le\alpha\big] }

\paragraph{Minimum Description Length}%
\index{MDL}%
\index{minimum description length|see{MDL}}
The Minimum Description Length or short MDL\margincite{MDL} of a
string~$s$ is the length of the shortest two-part code for~$s$ that
uses less than $\alpha$ bits. It consists of the number of bits needed
to encode the model $m_p$ that defines a distribution and the negative
log likelihood of~$s$ under this
distribution.\margincite{$\lambda_s(\cdot)$}\index{l@$\lambda_s(\cdot)$}

\form{
  \lambda_s(\alpha) \eq \operatornamewithlimits{min}_{m_p}\big[
             -\log p(s|m_p) + K(m_p):
             p (s|m_p)>0,\;K(m_p)\le\alpha]
}

\pa

It has recently been shown by Nikolai Vereshchagin and Paul Vit\'anyi
in~\cite{Vitanyi:2002} that a model that minimizes the description
length also minimizes the randomness deficiency, though the reverse
may not be true. The most fundamental result of that paper is the
equality

\form{
    \beta_s(\alpha) 
  \eq 
    h_s(\alpha)+\alpha-K(s) 
  \eq
    \lambda_s(\alpha)-K(s)
}

\formskip

where the mutual relations between the Kolmogorov structure function,
the minimum randomness deficiency and the minimum description length
are pinned down, up to logarithmic additive terms in argument and
value.

\boldframe{

MDL minimizes randomness deficiency.  With this important result
established, we are very keen to learn whether MDL can minimize the
generalization error as well.

}

\subsection{Practical MDL}\label{mdl}\index{MDL|(}

\index{MDL!two-part MDL!definition}%
\index{Rissanen, J.|(}\index{penalty}
From 1978 on Jorma Rissanen developed the idea to minimize the
generalization error of a model by penalizing it according to its
description length~\cite{Rissanen:1978}. At that time the only other
method that successfully prevented overfitting by penalization was
the\index{Akaike}%
\index{AIC|see{Akaike information criterion}}%
\index{Akaike information criterion}
Akaike Information Criterion (AIC). The AIC selects the model $m_k$
according
to\margincite{$\learn_{AIC}(\cdot)$}\index{L_{AIC}@$\learn_{AIC}(\cdot)$}

\form{
  \learn_{AIC}(s) \eq \operatornamewithlimits{min}_k 
  \left[n\;\log\sigma_k^2+2k\right]
}

\formskip

where $\sigma_k^2$ is the mean squared error of the model $m_k$ on the
training sample $s$, $n$ the size of $s$ and $k$ the number of
parameters used. H. Akaike introduced the term $2k$ in his 1973
paper~\cite{Akaike:1973} as a penalty on the complexity of the model.

\pa

Compare this to Rissanen's original MDL
criterion:\margincite{$\learn_{Ris}(\cdot)$}\index{L_{Ris}@$\learn_{Ris}(\cdot)$}

\form{
 \learn_{Ris}(s) \eq \operatornamewithlimits{min}_k
 \left[\; -\log p(s|m_k)+k\log\sqrt{n}\;\;\right]
}

\pa

Rissanen replaced Akaike's modified error $n\log(\sigma_k^2)$ by the
information theoretically more correct term $-\log p(s|m_k)$.  This is
the length of the Shannon-Fano code for $s$ which is a good
approximation of $K(s|m_k)$, the complexity of the data given the
$k$-parameter distribution model $m_k$, typicality assumed\foot{%
For this approximation to hold, $s$ has to be typical for the model
$m_k$. See Section~\reference{min-ran-def} for a discussion of
typicality and minimum randomness deficiency.}.
Further, he penalized the model complexity not only according to
the number of parameters but according to both parameters and
precision.  Since statisticians at that time treated parameters
usually as of infinite precision he had to come up with a reasonable
figure for the precision any given model needed and postulated it to
be $\log\sqrt{n}$ per parameter. This was quite a bold assumption but
it showed reasonable results. He now weighted the complexity of the
encoded data against the complexity of the model. The result he
rightly called Minimum Description Length because the winning model
was the one with the lowest combined complexity or description length.
  
\pa

Rissanen's use of model complexity to minimize the generalization error
comes very close to what Ray Solomonoff\index{Solomonoff, R.} 
originally had in mind when he first developed \kc.  The maximum a
posteriori model according to Bayes' rule\index{Bayes' rule}, supplied
with Solomonoff's universal distribution\index{universal distribution},
will favor the Minimum Description Length model, since

\formsplit{
  \operatornamewithlimits{max}_m\;\left[P(m|s)\right]
\fsequal
  \operatornamewithlimits{max}_m\;\left[\frac{P(s|m)\;P(m)}{P(s)}\right]
\go\fsequal
  \operatornamewithlimits{max}_m\;\left[P(s|m)\;2^{-K(m)}\right]
\go\fsequal
  \operatornamewithlimits{min}_m\;\big[-\log P(s|m)+K(m)\big]
}

\pa

Though Rissanen's simple approximation of $K(m)\approx k\log\sqrt{n}$
could compete with the AIC in minimizing the generalization error, the
results on small samples were rather poor. But especially the small
samples are the ones which are most in need of a reliable method to
minimize the generalization error. Most methods converge with the
optimum results as the sample size grows, mainly due to the law of
large numbers which forces the statistics of a sample to converge with
the statistics of the source.  But small samples can have very
different statistics and the big problem of model selection is to
estimate how far they can be trusted.

%

\pa

In general, two-part MDL makes a strict distinction between the
theoretical complexity of a model and the length of the implementation
actually used. All versions of two-part MDL follow a three stage
approach:

\enums{ 

   \item the complexity $-\log p(s|m_k)$ of the sample according to
         each model $m_k$ is calculated at a high precision of $m_k$.

   \item the minimum complexity $K(m_k)$ which would theoretically be
         needed to achieve this likelihood is estimated.

   \item this theoretical estimate $E\big[K(m_k)\big]$ minus the
         previous $\log p(s|m_k)$ approximates the overall complexity
         of data and model.

}

\index{Rissanen, J.|)}

\paragraph{Mixture MDL}\index{MDL!mixture MDL!definition}
More recent versions of MDL look deeper into the complexity of the
model involved.  Solomonoff and Rissanen in their original approaches
minimized a two-part code, one code for the model and one code for the
sample given the model.  Mixture MDL leaves this approach.  We do no
longer search for a particular model but for the number of parameters
$k$ that minimizes the total code length $-\log p(s|k)+\log(k)$. To do
this, we average $-\log\, p(s|m_k)$ over all possible models $m_k$ for
every number of parameters $k$, as will be defined further
below.\margincite{$\learn_{mix}(\cdot)$}\index{L_{mix}@$\learn_{mix}(\cdot)$}

\form{
 \learn_{mix}(s) \eq \operatornamewithlimits{min}_k
 \Big[ -\log p(s|k)+\log k\Big]
}

\pa

Since the model complexity is reduced to $\log k$ which is almost
constant and has little influence on the results, it is not
appropriate anymore to speak of a mixture code as a two-part code.

\pa

Let $M_k$ be the $k$-dimensional parameter space of a given family of
models and let $p(M_k = m_k)$ be a \index{distribution!prior} prior
distribution over the models in $M_k$\foot{%
For the moment, treat models as vectors in $\mathbb{R}^k$ so that
integration is possible.  See the discussion on Fisher information in
Section~\reference{fisher} for a similar problem.}.
Provided this prior distribution is defined in a proper way we can
calculate the probability that the data was generated by a
$k$-parameter model as

\form{
  p(s|k) \eq \int_{m_k\in M_k}\; p(m_k)\;p(s|m_k)\;d m_k
}

\pa

Once the best number of parameters $k$ is found we calculate our model
$m_k$ in the conventional way. This approach is not without problems
and the various versions of mixture MDL differ in how they address
them:

\items{

  \item The binary models $m_k$ form only a discrete subset of the
        continuous parameter space $M_k$. How are they distributed
        over this parameter space and how does this effect the
        results?

  \item what is a reasonable prior distribution over $M_k$?

  \item for most priors the integral goes to zero or infinity. How do
        we normalize it?

  \item the calculations become too complex to be carried out in
        practice.

}

\paragraph{Minimax MDL}\index{MDL!mixture MDL!definition}
Another important extension of MDL is the minimax strategy.  Let $m_k$
be the $k$-parameter model that can best predict $n$ future values
from some i.i.d.\@ training values. Because $m_k$ is unknown, every
model $\hat{m}_k$ that achieves a least square error on the training
values will inflict an extra cost when predicting the $n$ future
values. This extra cost is the Kullback Leibler
distance\index{Kullback Leibler distance}

\form{
  D(m_k||\hat{m}_k)
    \eq
  \sum_{x^n\in X^n} p(x^n|m_k)\log\frac{p(x^n|m_k)}{p(x^n|\hat{m}_k)}.
}

\pa

The minimax strategy favors the model $m_k$ that minimizes the maximum
of this extra cost.\margincite{$\learn_{mm}(\cdot)$}

\form{
  \learn_\text{mm} \eq  
  \operatornamewithlimits{min}_{k} \;
  \operatornamewithlimits{max}_{m_k\in M_k} \;
  D(m_k||\hat{m}_k)
}

\index{MDL|)}

\subsection{Error minimization}

\index{Gauss, C.F.}%
Any discussion of information theory and complexity would be
incomplete without mentioning the work of Carl Friedrich Gauss
(1777--1855).  Working on astronomy and geodesy, Gauss spend a great
amount of research on how to extract accurate information from
physical measurements. Our modern ideas of error minimization are
largely due to his work.

\paragraph{Euclidean distance and mean squared error}%
\index{Euclidean distance}%
\index{mean squared error}
To indicate how well a particular function $f(x)$ can approximate
another function $g(x)$ we use the Euclidean distance or the mean
squared error. Minimizing one of them will minimize the other so which
one is used is a matter of convenience. We use the mean squared
error. For the interval $x\in[a,b]$ it is defined as

\dontsplit{

\form{
  \sigma^2_f \eq \frac{1}{b-a}\;\int_a^b\big( f(x)-g(x)\big)^2 \;dx
}

\pa

This formula can be extended to multi-dimensional space. 

}

\pa
\index{distribution!Gaussian}
Often the function that we want to approximate is unknown to us and is
only witnessed by a sample that is virtually always polluted by some
noise. This noise\index{noise} includes measurement noise, rounding
errors\index{rounding} and disturbances during the execution of the
original function. When noise is involved it is more difficult to
approximate the original function. The model has to take account of
the distribution of the noise as well. To our great convenience a mean
squared error\index{squared error} $\sigma^2$ can also be interpreted
as the variance of a Gaussian or normal distribution.  The Gaussian
distribution is a very common distribution in nature. It is also akin
to the concept of Euclidean distance, bridging the gap between
statistics and geometry.  For sufficiently many points drawn from the
distribution
$\gaussian\big(f(x),\,\sigma^2\big)$\margincite{$\gaussian(\cdot,\cdot)$}
the mean squared error between these points and $f(x)$ will approach
$\sigma^2$ and approximating a function that is witnessed by a sample
polluted by Gaussian noise becomes the same as approximating the
function itself.

\pa

\newcommand{\myDistance}{\ensuremath{l\hspace{.3mm}}}

Let $a$ and $b$ be two points and let $\myDistance$ be the Euclidean distance
between them.  A Gaussian distribution $p(\myDistance)$ around $a$ will assign
the maximum probability to $b$ if the distribution has a variance that
is equal to $\myDistance^2$.  To prove this, we take the first derivative of
$p(\myDistance)$ and equal it to zero:

\formsplit{
  \frac{\operatorname{d}}{\operatorname{d}\sigma}\;p(\myDistance)
\fsequal
  \frac{\operatorname{d}}{\operatorname{d}\sigma} \;
  \frac{1}{\sigma\sqrt{2\pi}}\;e^{-\myDistance^2/2\sigma^2}
\go\fsequal
    \frac{-1}{\sigma^2\sqrt{2\pi}}\;e^{-\myDistance^2/2\sigma^2}
  + \frac{1}{\sigma\sqrt{2\pi}}\;e^{-\myDistance^2/2\sigma^2}
    \left(\frac{2\,\myDistance^2}{2\,\sigma^3}\right)
\go\fsequal
    \frac{1}{\sigma^2\sqrt{2\pi}}\;e^{-\myDistance^2/2\sigma^2}
    \left(\frac{\myDistance^2}{\sigma^2} - 1\right)
\go\fsequal
    0
}

\formskip

which leaves us with

\form{
  \sigma^2 \eq \myDistance^2 . 
}

\pa

Selecting the function $f$ that minimizes the Euclidean
distance between $f(x)$ and $g(x)$ over the interval $[a,b]$ is the
same as selecting the maximum likelihood distribution\index{maximum
likelihood}, the distribution $\gaussian\big(f(x),\,\sigma^2\big)$
that gives the highest probability to the values of $g(x)$.

\paragraph{Maximum entropy distribution}
Of particular interest to us is the entropy of the Gaussian
distribution. The optimal code for a value that was drawn according to
a Gaussian distribution $p(x)$ with variance $\sigma^2$ has a mean
code length or entropy of

\formsplit{
  H(P) 
\fsequal
  -\int_{-\infty}^{\infty}p(x)\log p(x)\;dx
\go\fsequal
 \frac{1}{2}\log\left(2\pi e \sigma^2\right)
}

\pa

To always assume a Gaussian distribution for the noise may draw some
criticism as the noise may actually have a very different
distribution. Here another advantage of the Gaussian distribution
comes in handy: for a given variance the Gaussian distribution is the
maximum entropy distribution. \index{maximum entropy} It gives the
lowest log likelihood to all its members of high probability. That
means that the Gaussian distribution is the safest assumption if the
true distribution is unknown. Even if it is plain wrong, it promises
the lowest cost of a wrong prediction regardless of the true
distribution~\cite{Grunwald:2000}.

\subsection{Critical points}
\label{critical}
\index{critical points} 

Concluding that a method does or does not select a model close to the
optimum is not enough for evaluating that method.  It may be that the
selected model is many degrees away from the real optimum but still
has a low generalization error.  Or it can be very close to the
optimum but only one degree away from overfitting, making it a risky
choice. A good method should have a low generalization error {\em
and\/} be a save distance away from the models that overfit.

\pa

How a method evaluates models other than the optimum model is also
important. To safeguard against any chance of overfitting we may want
to be on the safe side and choose the lowest degree model that has an
acceptable generalization error. This requires an accurate estimate of
the per model performance.  We may also combine several methods for
model selection and select the model that is best on average. This too
requires reliable per model estimates.  And not only the
generalization error can play a role in model selection. Speed of
computation and memory consumption may also constrain the model
complexity. To calculate the best trade off between algorithmic
advantages and generalization error we also need accurate per model
performance.

\pa

When looking at the example we observe a number of critical points
that can help us to evaluate a method:

\descriptions{
 
  \item[the origin:]\index{origin} the generalization error when
        choosing the simplest model possible. For
        polynomials this is the expected mean of $y$ ignoring $x$.

  \item[the initial region:]\index{initial region} may
        contain a local maximum slightly worse than the origin or a
        plateau where the generalization error is almost constant.

  \item[the region of good
        generalization:]\index{generalization!region of good} the
        region that surrounds the optimum and where models perform
        better than half way between origin and optimum.
        Often the region of good generalization is visible in the
        generalization analysis as a basin with sharp increase and
        decrease at its borders and a flat plateau in the center where
        a number of competing minima are located.

  \item[the optimum model:]\index{model!optimum} the minimum within
        the region of good generalization.

  \item[false minima:]\index{false minima} models that show a single
        low generalization error but lie outside or at the very edge
        of the region of good generalization.

  \item[overfitting:]\index{overfitting!point of real} from a certain
        number of degrees on all models have a generalization error
        worse than the origin. 

}

Let us give some more details about three important features:

\paragraph{Region of good generalization}
\index{generalization!region of good}
The definition of the region of good generalization as better than
half way between origin and optimum needs some explanation. Taking
an absolute measure is useless as the error can be of any magnitude. A
relative measure is also useless because while in one case origin and
optimum differ only by 5 percent with many models in between, in
another case even the immediate neighbors might show an error two
times worse than the optimum but still much better than the origin.

\pa

Better than half way between origin and optimum may seem a rather weak
constraint.  With big enough samples all methods might eventually
agree on this region and it may become obsolete. But we are primarily
concerned with small samples. And as a rule of thumb, a method that
cannot fulfill a weak constraint will be bad on stronger constraints as
well.

\paragraph{False minima}\index{false minima}
Another point that deserves attention are the false minima. 
While different samples of the same size will generally agree on the
generalization error at the origin, the initial region, the region of
good generalization and the region of real overfitting, the false
minima will change place, disappear and pop up again at varying
amplitudes. They can even outperform the real optimum. The reason for
this can lie in rounding errors during calculations and in the random
selection of points for the training set.  And even though the
training sample might miss important features of the source due to its
restricted size, the model might hit on them by sheer luck, thus
producing an exceptional minimum.

\pa

Cross-validation particularly suffers from false minima and has to be
smoothed before being useful. Taking the mean over three adjacent
values has shown to be a sufficient cure. Both versions of MDL seem to
be rather free of it.

\paragraph{Point of real overfitting}
\index{overfitting!point of real}
The point where overfitting starts also needs some explanation.  It is
tempting to define every model larger than the optimum model as
overfitted and indeed, this is often done. But such a definition
creates a number of problems. First, the global optimum is often
contained in a large basin with several other local optima of almost
equal generalization error. Although we assume that the training
sample carries reliable information on the general outline of
generalization error, we have no evidence that it carries information
on the exact quality of each individual model. It would be wrong to
judge a method as overfitting because it selected a model 20 degrees
too high but of a low generalization error if we have no indication
that the training sample actually contained that information. On the
other hand, at the point on the $x$-axis where the generalization
becomes forever worse than at the origin the generalization error
usually shows a sharp increase. From this point on differences are
measured in orders of magnitude and not in percent which makes
it a much clearer boundary. Also, if smoothing is applied, different
forms of smoothing will favor different models within the region of
good generalization while they have little effect on the location of
the point where the generalization error gets forever worse. And
finally, even if a method systematically misses the real optimum, as
long as it consistently selects a model well within the region of good
generalization of the generalization analysis it will lead to good
results. But selecting or even encouraging a model beyond the point
where the error gets forever worse than at the origin is definitely
unacceptable.

\bibliographystyle{alpha}


\end{document}